# Revealing Hidden Potentials of the q-Space Signal in Breast Cancer

Paul Jaeger[1], Sebastian Bickelhaupt[2], Frederik Bernd Laun[2,3], Wolfgang Lederer[4], Daniel Heidi[5], Tristan Anselm Kuder[6], Daniel Paech[2], David Bonekamp[2], Alexander Radbruch[2], Stefan Delorme[2], Heinz-Peter Schlemmer[2], Franziska Steudle[2], and Klaus H. Maier-Hein[1]

[1] Medical Image Computing, German Cancer Research Center (DKFZ), Heidelberg, Germany
[2] Department of Raiology, DKFZ, Heidelberg, Germany
[3] Institute of Radiology, University Hospital Erlangen, Germany
[4] Radiological Practice at the ATOS Clinic, Heidelberg, Germany
[5] Radiology Center Mannheim (RZM), Germany
[6] Medical Physics in Radiology, DKFZ, Heidelberg, Germany

**Abstract.** Mammography screening for early detection of breast lesions currently suffers from high amounts of false positive findings, which result in unnecessary invasive biopsies. Diffusion-weighted MR images (DWI) can help to reduce many of these false-positive findings prior to biopsy. Current approaches estimate tissue properties by means of quantitative parameters taken from generative, biophysical models fit to the q-space encoded signal under certain assumptions regarding noise and spatial homogeneity. This process is prone to fitting instability and partial information loss due to model simplicity. We reveal unexplored potentials of the signal by integrating all data processing components into a convolutional neural network (CNN) architecture that is designed to propagate clinical target information down to the raw input images. This approach enables simultaneous and target-specific optimization of image normalization, signal exploitation, global representation learning and classification. Using a multicentric data set of 222 patients, we demonstrate that our approach significantly improves clinical decision making with respect to the current state of the art.

## 1 Introduction

Breast cancer is the most frequently diagnosed cancer type among women [1]. While it has been shown that X-ray mammography decreases breast cancer-related mortality, it suffers from high amounts of false positive findings, which lead to overdiagnosis of clinically insignificant lesions [2]. Additional non-invasive examination based on DWI was recently proposed as a powerful yet light-weight addition to the screening process [3]. In DWI, the signal behavior at different diffusion gradients is quantitatively characterized by fitting biophysical models to the signal and inferring apparent tissue properties from them. The state of the art method in breast cancer DWI is Diffusion Kurtosis Imaging (DKI), where the apparent diffusion coefficient (ADC) and the apparent kurtosis coefficient (AKC) are extracted representing Gaussian and non-Gaussian diffusion, respectively [4, 5]. Using DKI, state of the art results for breast lesion classification have been reported recently [6, 7]. Model-based methods, however, are simplified approaches to physical processes, making them prone to partial information loss and dependent on explicit prior physical knowledge resulting in potential fitting instabilities and limited generalization abilities. These shortcomings have led to an emergence of a broad spectrum of signal and noise models designed under different assumptions. Recent studies in brain imaging have shown how deep learning can circumvent some of the disadvantages related to classical model-based approaches in diffusion MRI data processing [8–11]. However, the currently existing learning-based approaches cannot be more knowledgeable than the classical model-based approach used as ground truth during training. Thus, the performance of existing model-free approaches is currently sealed, and the main benefit so far was found in the reduction of requirements on the input data side, e.g. saving acquisition time.

In this paper we show in a first clinical scenario how model-free diffusion MRI can be integrated into an end-to-end training, thus directly relating clinical information to the raw input signal. By backpropagating this information through an integrative CNN architecture, simultaneous and target-specific optimization of image normalization, signal exploitation, global representation learning and classification is achieved. We demonstrate the superiority of our approach for clinical decision making by performing breast lesion classification on a multicentral data set of 222 patients with suspicious findings in X-ray mammography.

## 2 Methods

**MRI Dataset** This study is performed on a combined data set of 222 patients acquired in two study sites with 1.5 T MR scanners from different vendors. Images were acquired with the b-values 0, 100, 750 and $1500 \, \mathrm{s \, mm^{-2}}$ and a slice thickness of 3 mm. The in-plane resolution of one scanner had to be upsampled by a factor 2 to match the other scanner's resolution of 1.25 mm. Figure 1 shows an example set of diffusion-weighted images for one patient. All patients were diagnosed with a BI-RADS score of 4 or higher in an earlier X-ray mammography. Following DWI, core-needle biopsy was performed identifying 122 malignant and 100 benign lesions. Manual segmentation of regions of interest (ROI) on the lesions was conducted by an expert radiologist without knowledge about the biopsy results. 23 of the lesions were not visible on the diffusion-weighted images and predicted as benign.

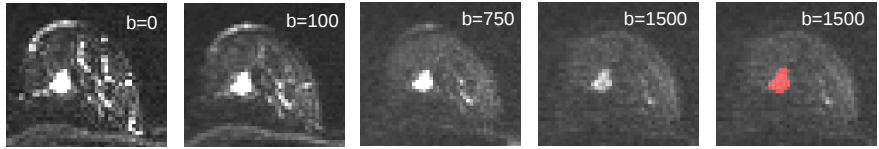

**Fig. 1.** Set of diffusion-weighted images at different b-values for one patient including an example segmentation of a malignant lesion on the $b = 1500\,\mathrm{s\,mm^{-2}}$ image.

**Diffusion Kurtosis Imaging** DKI is the current state of the art for DWI in breast cancer. In DKI, the ADC and the AKC indicate presence of diffusion restricting barriers and tissue heterogeneity. The two coefficients are derived for every voxel in the ROI by fitting its signal intensities $S(b)$ to:

$$S(b) = S_0 \ \exp\left(-b\ \mathrm{ADC} + \frac{1}{6}\ b^2 \mathrm{ADC}^2 \mathrm{AKC}\right), \qquad (1)$$

where $S_0 = S(b=0)$ and $b$ is the DWI b-value [4]. The global coefficients of a lesion are determined by averaging the coefficients of all voxels in the lesion [6].

In this study the signal intensity $S_0$ was omitted during the fit due to its instability. It was instead used as a third free parameter initialized by the measured value [12]. The resulting coefficients ADC and AKC were required to lie within $0 < \mathrm{ADC} < 3.5\,\mathrm{\mu m^2\,ms^{-1}}$ and $0 < \mathrm{AKC} < 3$, i.e. voxels yielding values outside these intervals were excluded from the averaging to decrease the influence of fitting instabilities.

**End-to-End q-space Deep Learning** The recently proposed q-space deep learning method [8] uses neural networks to imitate model-based approaches like DKI by training them on model-derived parameters. In contrast to that method, we aim to replace the model-based approaches by not using any model-related parameters as training target. Instead we train our approach directly on targeted clinical decision. By integrating the entire data processing pipeline into a CNN and training it end-to-end, this valuable information is backpropagated through the network optimizing all pipeline components on the specific clinical task. This enables our approach to yield performances beyond model-based methods. The proposed architecture consists of four modules:

*Input and Image Normalization Module.* The proposed CNN architecture is developed to operate directly on the diffusion-weighted images as input, where each of the four b-value images is assigned to a corresponding input channel of the network. For the task of lesion classification every image is cropped to a bounding box around the segmented ROI and voxels outside of the ROI are set to 0. Image normalization can be essential when working with raw MRI signal intensities. To facilitate this step, we measure the mean signal intensity of an additional ROI placed in a fat area of a breast in each image. The measured value is arrayed to match the shape of the corresponding lesion ROI and provided to the CNN as a fat intensity map in an additional input channel.

*Signal Exploitation Module.* The input is processed by layers of 1x1 convolutions, which only convolve the signals in each separate voxel across the input channels. This method is equivalent to applying a multilayer perceptron to each voxel, like it is done in q-space Deep Learning, i.e it enables the network to exploit the information contained in the differently weighted signals for a voxel. The additional input channel for image normalization extends the set of differently weighted signals in every voxel by the corresponding value of the fat intensity map, thus transferring normalizing information about the image into all 1x1 convolutions. Three layers of 1x1 convolutions are applied transforming the input data into 512 feature maps. In analogy to model-based diffusion coefficients, we term these representations *deep diffusion coefficients* (DDC), where each of the maps corresponds to one coefficient.

*Global Representation Learning Module.* Learning global representations, e.g. for texture or geometric analysis, requires inter-voxel convolutions. To this end, the DDC maps are processed by two blocks of three 3x3 convolutional layers, while downsampling the input sizes between the blocks using 2x2 max pooling. In principal, this component can be repeated arbitrarily, but is limited in the case of lesion classification by the small input sizes of the lesion ROIs.

*Binary Classification Module.* The final convolutional layer containing the learned global representations in form of feature maps is followed by a global average pooling layer, which aggregates the representations by transforming each feature map into a single mean value. Note, that this all convolutional architecture allows for variable input sizes by avoiding any classical dense layers, which we exploit by processing ROIs of different shapes through the same network [13]. The output is a vector with the length of the number of feature maps containing the global representations. This feature vector is used as input for a softmax layer transforming the features into class probabilities, which the binary classification is performed on using a categorical cross entropy loss function. By training the proposed network architecture in an end-to-end fashion, image normalization, signal exploitation and global representations are learned simultaneously and optimized directly for the classification problem.

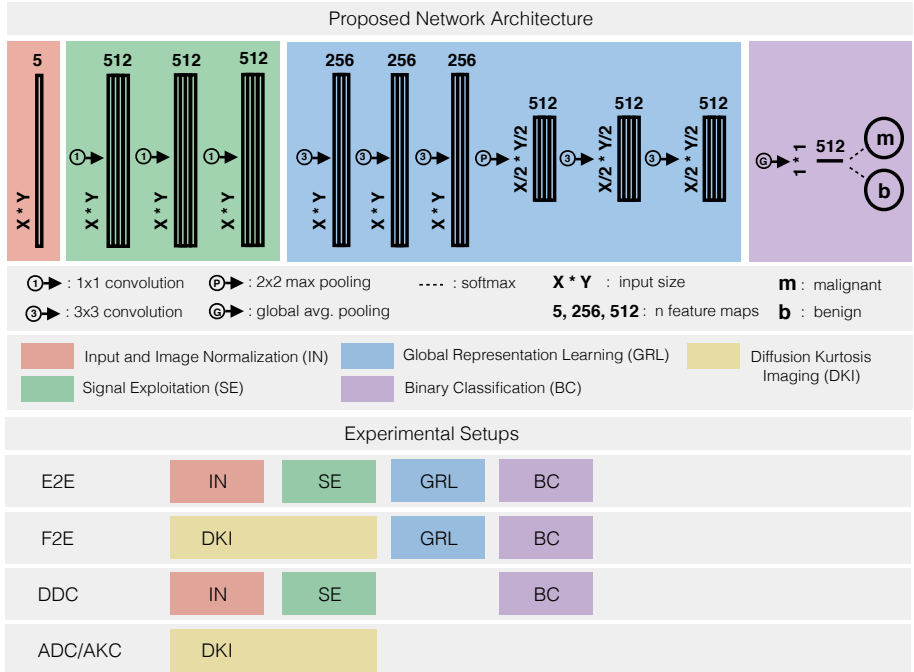

**Fig. 2.** Detailed network architecture for the different experimental setups explored in this paper. All convolutional layers use ReLU activation functions. The 5 input channels receive the diffusion-weighted images at four b-values plus the fat intensity map.

**Experimental Setups** In order to investigate the potential improvement of the proposed approach for clinical decision making, a two step evaluation is performed. First, the model-free signal exploitation is assessed by comparing the classification performances of the DDC and the means of the model-based coefficients ADC and AKC. A mean DDC is generated by applying the global average pooling directly to the DDC feature maps. In a second step, the end-to-end (E2E) approach, and the simultaneous optimization of all DWI data processing components in particular, is evaluated by benchmarking it against a model-based CNN method. For this benchmark, the E2E architecture is modified by feeding parametric maps of ADC and AKC into the global representation module instead of the self-learned DDC feature maps. For simplicity, the benchmark method is referred to as the fit-to-end method (F2E). All experimental setups and the detailed network architecture are shown in Figure 2.

**Training Details** Experiments were run using 10-fold cross validation (CV) with 80% training data, 10% validation data, and 10% test data, where the validation data was used for hyper parameter search and the corresponding loss as a stopping criterion. Batches were generated by randomly sampling 25 slices of each target class. Notably, all slice samples are mutually exclusive w.r.t single patients to provide strict data separation. The size of the input layer was chosen according to the size of the largest lesion in the batch, while zero padding the smaller lesions. Note, that this method results in batches of variable shape, which is accounted for by the model's all convolutional architecture. All images were masked according to the segmented lesion, i.e. voxels outside the ROI were set to 0. Data augmentation was performed batch-wise by randomly mirroring left-right, up-down or rotating by 90°. Dropout was applied to all convolutional layers with $p = 0.5$. The learning rate was initialized at $lr = 0.0005$ and decreased each epoch by a factor 0.985. The model was trained using categorical cross-entropy loss over 12 epochs, processing 100 batches per epoch. Inference was done by processing each slice $j$ of a patient $i$ individually and weighting the obtained predictions $p_{i,j}$ with the number of voxels $v_{i,j}$ in the slice against the overall number of voxels in the lesion $v$ in order to obtain the prediction $p_i$ for a patient:

$$p_i = \frac{1}{v} \sum_{j=1}^{s} p_{i,j} * v_{i,j} \qquad (2)$$

An ensemble of fifteen classifiers was trained for each fold of the CV and the resulting $p_i$ for one patient were averaged for the final ensemble prediction.

**Statistical Evaluation** Models were compared by evaluating the accuracy score on the test set. The decision threshold $t_c$ was chosen at sensitivity $se \geq 0.96$. This relatively high threshold matches the sensitivity of core-needle biopsy as reported in literature [14], thus ensuring the integrative character of DWI as a follow-up study of mammography. The resulting specificities at $t_c$, i.e. the percentage of removed false positives from the test set, were tested for significance using the McNemar-Test (significance level $\alpha = 0.05$). Note, that statistics were calculated across all CV folds, i.e. test set predictions of each fold were collected and fused to a final test set containing all patients.

## 3 Results

Table 1 shows a comparison amongst all methods explored in this paper. On the studied data set with 100 benign lesions (false positive mammographic findings) and 122 malignant lesions (true positive mammographic findings), the previously chosen decision threshold yields a sensitivity of 0.967 for all methods. This corresponds to correctly identifying 118 out of the 122 true positives. The E2E approach shows best performances with an accuracy of $0.815 \pm 0.026$ and a specificity at $t_c$ of $0.630 \pm 0.048$, correctly identifying 63 of the 100 false-positives. This significantly (p-value $< 0.01$) improves the clinical decisions with respect to the F2E method, which has an accuracy of $0.743 \pm 0.029$ and a specificity at $t_c$ of 0.470 ± 0.050, correctly identifying 47 of the 100 false-positives. Comparing classification performances of the coefficients without additional global representation learning, the DDC shows the highest accuracy of 0.770 ± 0.028 and the highest specificity at $t_c$ of 0.530 ± 0.050 outperforming the model-based coefficients, which both have an accuracy of 0.734 ± 0.030 and a specificity at $t_c$ of 0.450 ± 0.050. As an additional threshold-independent analysis, the area under the receiver operator curve (AUC) was studied. Here, the E2E approach is also superior with an AUC 0.907 ± 0.038 compared to 0.886 ± 0.043 for F2E. Among the explored coefficients DDC shows the best AUC with 0.868 ± 0.043 compared to 0.827 ± 0.056 for ADC and 0.799 ± 0.056 for AKC.

**Table 1.** Results on the test data of all methods explored in this paper.

| Method | AUC | Acc. at $t_c$ | Spec. (Sens.) at $t_c$ | $t_c$ |
|---|---|---|---|---|
| E2E | **0.907** ± 0.038 | **0.815** ± 0.026 | **0.630** ± 0.048 (0.967) | ≥ 0.418 |
| F2E | 0.886 ± 0.043 | 0.743 ± 0.029 | 0.470 ± 0.050 (0.967) | ≥ 0.34 |
| DDC | **0.868** ± 0.043 | **0.770** ± 0.028 | **0.530** ± 0.050 (0.967) | ≥ 0.29 |
| ADC | 0.827 ± 0.056 | 0.734 ± 0.030 | 0.450 ± 0.050 (0.967) | ≤ 1.83 |
| AKC | 0.799 ± 0.056 | 0.734 ± 0.030 | 0.450 ± 0.050 (0.967) | ≥ 0.845 |

## 4 Discussion

The results show that our approach significantly improves clinical decision making compared to the current state of the art. We first demonstrate, how data-driven signal exploitation in DWI outperforms the current model-based methods and show in a second step how this approach can be integrated into and end-to-end CNN architecture. On our data set, the end-to-end training is able to prevent an additional 16 out of 100 women from overdiagnosis with respect to the benchmark method using model-based coefficients as input. This benchmark is designed in such a way, that credits for improvement of our approach can be clearly assigned to the data-driven signal exploitation and its integrability to joint optimization. In contrast to recent data-driven methods like q-space deep learning, which are trained on model-related parameters, our end-to-end training is trained directly on the targeted clinical decision. This enables our approach to optimize all components of the data processing pipeline simultaneously on a specific task, thus not being limited by model assumptions. A limitation to our approach is the dependence on manual segmentation of lesions, which can be addressed in future studies by integrating automated segmentation into the network architecture. The dependence on specific b-values as network inputs is a further limitation, which can be tackled by means like domain adaption [15]. Also we increasingly observe efforts towards standardization of DWI protocols. The multicentric character of the utilized data set hints upon the generalization and normalization abilities of the method across different input characteristics.